\documentclass[review]{elsarticle}
\usepackage{lineno,hyperref}
\usepackage{algorithmic}
\usepackage{algorithm}
\usepackage[table,xcdraw]{xcolor}
\usepackage{graphicx}
\def\rotatecharone#1{\rotatebox[origin=c]{180}{#1}}
\usepackage{booktabs}
\usepackage{longtable}
\usepackage{amsmath}
\graphicspath{ {fig/} }
\usepackage{tabularx}
\modulolinenumbers[5]
\journal{Journal of Pattern Recognition}

\begin{document}

\begin{frontmatter}

\title{Dataset Structural Index: Leveraging a machine’s perspective towards visual data}


\author[1]{Dishant Parikh\corref{cor1}}
\ead{dishant30899@gmail.com}
\cortext[cor1]{Corresponding author}
\address[1]{Independent Researcher.}

\begin{abstract}
With advances in vision and perception architectures, we have realized that working with data is equally crucial, if not more, than the algorithms. Till today, we have trained machines based on our knowledge and perspective of the world. The entire concept of Dataset Structural Index(DSI) revolves around understanding a machine’s perspective of the dataset. With DSI, I show two meta values with which we can get more information over a visual dataset and use it to optimize data, create better architectures, and have an ability to guess which model would work best. These two values are the Variety contribution ratio and Similarity matrix. In the paper, I show many applications of DSI, one of which is how the same level of accuracy can be achieved with the same model architectures trained over less amount of data. 

\end{abstract}


\end{frontmatter}

\section{Impact statement}

DSI helps in making better decisions about the dataset choice and the model architecture to be used. DSI, when used effectively, can be a powerful tool in investigating the dataset and optimize it. DSI could be a new standard for visual datasets and a step further into data-centric approaches. The best thing about DSI is the simplicity of usage and the amount of information extracted from just two values for each class.

Dataset structural index provides four critical capabilities:

\begin{enumerate}
    \item To effectively assess the dataset before detailed analysis. 
    \item To move around quickly from one dataset to another by getting a better understanding as to which one to explore more, based on the application and resources available.
    \item To efficiently optimize a visual dataset according to the task at hand. 
    \item To ease the decision-making process by giving a logical guess as to which model would work best, given the DSI values.
\end{enumerate}

\section{Introduction}
When statistical machine learning was introduced, a lot of work was done in developing strategies to understand a particular dataset. But when it came to visual datasets, the field immediately stepped towards the algorithmic side. One of the fundamental reasons was the amount of information needed to translate from an image. But with the introduction of convolutional networks and transfer learning \cite{sharif2014cnn}, \cite{lecun1995convolutional}, \cite{shin2016deep}, it is possible to convert an image or a visual object into feature vectors without losing too much information about the entity under concern. It defined a way to use feature maps to compare and distinguish one visual object from another \cite{jmour2018convolutional}.

There has been a lot of work in using these feature vector conversions in systems like content-based image retrievals \cite{gudivada1995content}, using feature vectors as representations of different scenarios \cite{datta2005content}, \cite{eakins1999content}. It is critical to understand that there is a difference between the way a machine looks at the data and the way we do. There are scenarios in which the interpretation through features is a little different from the interpretation of humans. DSI is there to bridge the gap and understand the machine’s perspective before molding it to shape better architectures, in turn, better model performances. I think two concepts could be linked together to understand a machine’s viewpoint while working with visual data. First, the evaluation of the dataset. According to the machine, how much work would be needed to effectively transfer the knowledge from the dataset into the model? 
And second, the amount of data ”actually” required to fulfill the learning of the model. Keeping this in mind, I define two fundamental concepts, Variety contribution ratio, and Similarity matrix.

\subsection{Definitions}
\begin{enumerate}
    \item Variety contribution ratio ($v_r \in R^{1*m}$) is defined as the ratio of total images contributing to the variety ($v$) of a class and the total number of samples ($n$) in the class.
    \item Similarity matrix ($S \in R^{m*m}$) is defined as the pairwise similarity calculated between each class inside the dataset ($H$) with $m$ classes.
\end{enumerate}

\subsection{Variety contribution ratio }

Before I discuss the variety contribution ratio ($v_r$) further, it is critical to discuss why we are considering the variety($v$) and what exactly do I mean by variety. So, the variety of a class is the number of samples that are correctly classified in the class, and are varied enough to make sure the model is getting all kinds of scenarios to learn from. The main reason why I am considering variety is that it is fundamentally the most important part when it comes to convolutional mechanisms, as we know that the more scenarios covered, the better the model’s learning \cite{barbedo2018impact}. That is why we need to know how many samples actually bring in variety rather than being a redundant data sample. The best way to know this is to compare the feature vectors because, in the end, we need to evaluate the dataset from a machine’s perspective. There are images that we think are too distinct, but machines may not agree.

\begin{figure}[!t]
\centering
\includegraphics[width = 0.9\columnwidth,scale=1.5]{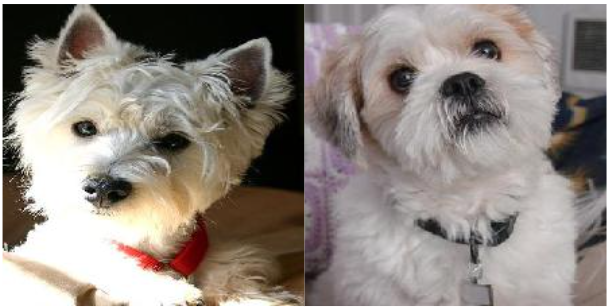}
\caption{Left: Scotch-Terrier and Right: Lhasa}
\label{figure1}
\end{figure}

Figure \ref{figure1} is an example, where two images look very similar but have the highest distance in their feature vectors (Cosine distance of 0.2665, which is very high). It proves that sometimes we may think that a particular data sample is just repetitive, while it may not be so. The same is true the other way around as well. We may think that the sample is quite distinct and important to teach the machine, but the distance might not be that high, and it may not contribute to the model’s learning. Hence, the variety contribution ratio is one of the two fundamental values that create the DSI. 

\subsection{Similarity matrix}

The variety contribution ratio ($v_r$) is an intracluster value of the DSI, i.e., it considers each class’s independent behavior. To have an intercluster value as well, I included the similarity matrix ($S$) in the DSI calculation. The similarity matrix takes into consideration the class-to-class relationship in DSI. It works similarly for contributing to a machine’s learning. Basically, in an image classification problem, the model’s training would be easy if the inter-class distance is relatively high. In the same way, if it is low, the work for the machine would be a little bit more complex. That is why knowledge of the distance between the classes is a crucial parameter of the DSI.

\section{Related works}
Since the introduction of deep convolutional neural networks[1], researchers have been relentlessly working on improving the visual perception of a machine. There have been tremendous amounts of work in terms of algorithmic design to extract finer and finer features from an image[3]. Starting with LeNet \cite{lecun1995convolutional} and AlexNet \cite{krizhevsky2012imagenet}, then VGG-16 \cite{simonyan2014very}, InceptionNet \cite{szegedy2015going}, to ResNet50 \cite{he2016deep} and ResNeXt50 \cite{xie2017aggregated}. Each of the algorithms gave a better accuracy than the previous algorithms and improved upon and gave birth to a lot more CNN-based architectures. But while the work on deeper networks was going on, there was one more contributing factor to the increase in the visual score of each new model - data. 

The field of deep learning has gotten massive attention due to the increase in data and computational resources available to train these algorithms. But unfortunately, the work done on optimizing and understanding datasets is still quite limited. There has been some work on the impact of dataset size on the model accuracies, notably by \cite{barbedo2018impact}. But the discussion can and should be extended to how much of that huge dataset “actually” contribute to the model’s training. It gave rise to the idea of a variety contribution ratio. Not only this but, one more criterion demands attention, the difficulty of classifying some classes from the others. 

When the field of machine vision was introduced the initial thesis was to get machines as close to humans as possible. But we have been keeping them apart until now. We have not taken into consideration the way humans confuse two relatively similar-looking things. If human beings find it difficult to classify two things properly even when we can get other things quite well, it raises a question that what if there are actual classes where machines might face a similar problem. It furthered the discussion on the similarity matrix. There has been a lot of work in the field of image similarity and the applications through it \cite{ding2020image}, \cite{ragkhitwetsagul2018picture}, \cite{plummer2020these}.

The work started in Image similarity quite early and it took off with the introduction of CBIR systems \cite{gudivada1995content} which are used quite heavily in many fields. The CBIR system introduced a way of using the features extracted from these state-of-the-art algorithms and using those to compare the images already in the dataset. It gave rise to using feature extractors in image similarity calculation. We have seen the effectiveness of correlation between the image feature histograms as the similarity measure \cite{wang2020image}. Not just the content-based similarity but many applications have considered the structural similarity \cite{shnain2017feature}. The same work was improved by C Zhang in PPIS-JOIN \cite{zhang2021ppis}. 

\section{Methodology}

\subsection{Dataset information}

\begin{enumerate}
    \item \textbf{Stanford dog breeds dataset}. 
    
    Stanford dog breeds dataset \cite{khosla2011novel} contains 20,580 images of 120 dog breeds from across the world. Each class contains roughly 150 images. I have used two versions of this dataset. The original one with 20,580 images, and the other, was optimized by the variety contribution ratio containing 19,093 images. Each image was cropped for reducing the noise before passing it to the training. Apart from that, the only other preprocessing was the general data augmentation and normalization appended with the PyTorch pipeline. 
    
    \item \textbf{Oxford Flowers dataset}. 
    
    Oxford Flowers dataset \cite{nilsback2008automated} contains 8,189 images of 102 flower species. Each class consists of between 40 and 258 images. I have used two versions of this dataset. The original one with 8,189 images, and the other, was optimized by the variety contribution ratio containing 5724 images. This particular dataset also is used to show various similarity matrix applications and how it can help in making a better choice of the model architecture. 
    
    \item \textbf{Four-class custom dataset (High similarity)}.
    
    The four-class dataset is a custom classification dataset collected and created using the classes from the COCO dataset \cite{lin2014microsoft} and some manual collection of images. The four classes are Bus, Car, Van, and Truck. There are a total of 1104 data samples in the dataset with roughly 250 images per class\footnote{This dataset can be found at-https://bit.ly/FourClassData}. The dataset gives insight into a small and highly similar dataset and helps in understanding the balance between the variety-contribution ratio and similarity matrix.

\end{enumerate}

\subsection{Significance of feature-based Image similarity}

Before I discuss the methodology, it is critical to answer the question, why only feature-based similarity is considered and not other types of similarity like SSIM(Structural similarity)? It is worth noting that the entire concept of DSI is in understanding the machine’s perspective of the dataset, not in getting the similarity as close as humans. The main difference between the previous approaches and DSI is that we are taking what the machine has seen about the object and treating that as a knowledge base, and getting the understanding from there. The point of the method is to not get the perfect similarity measure but to understand what the machine sees and leverage that knowledge for bettering our architectures and pipelines. 

\subsection{Feature Extraction}

The first step into the generation of the DSI is in understanding the machine’s perspective, .i.e. features that the machine extracts. For the same, I use a pretrained ResNet152 and make a forward pass on all the images. The feature vector ($X_i \in R^{1*2048}$) is extracted by taking the vector ($v_{ap}$) before it is passed to the average pooling layer ($f_{AP}$). The reason for extracting the features from the last layer is because the latter layers have more specific features while the previous ones have more general features \cite{yosinski2014transferable}. For feature extraction, there are two other layers available one is passing the vector to max-pooling and then extracting it. The second is to take the vector ($v_{fc} \in R^{1*1000}$) at the fully connected layer. But according to the work by \cite{Pohchih2020}, the features work well in the case of Avg. pooling especially while working with content-based image similarity. Hence, the feature matrix ($X \in R^{n*2048}$) for one class ($c$) is calculated by:
\begin{equation}
    X_{i,:} = f_{ap}(H_c,i)
\end{equation}

Another case to note is the model selection. Although any model can be used for feature extraction, the choice will depend a lot on the depth of feature extraction required. I have noticed that in most cases, the specific features are better extracted with deeper extraction. The argument can also be supported with the results on classification algorithms comparison \cite{bressem2020comparing} as well as the comparative results in the model choice in CBIR systems \cite{Pohchih2020}.

\subsection{Similarity matrix generation}

There are three steps to similarity matrix generation. 
\begin{enumerate}
    \item Feature centroid matrix ($T \in R^{m*2048}$) generation for the dataset.
    \item Calculating similarity ($s_{i,j}$) by comparing the distances of feature centroids ($T_i$ and $T_j$).
    \item Generating similarity matrix ($S \in R^{m*m}$) by placing the pairwise distance evaluations of the similarity for each class in the dataset.
\end{enumerate}

Feature centroid $T_i$ is generated by fitting an unsupervised clustering algorithm ($f_{cluster}$) with hyperparameter $n_{cluster} = 1$ and extracting the cluster center as the feature centroid. The use of clustering algorithms also helps in checking the inertia, that is the overall spread (Interia $q$) of the features in the class, which in turn can be used for checking the overall variety ($v$) of the class. But this way of checking variety works only if there is no other factor to be taken into account. This point is discussed further in Section \ref{variety}. There are two reasons for computing $T$: One, reducing the number of comparisons required. Second, the class should not be checked based on either outliers or extremes that will only hurt the distance computations for $S$. 

\begin{equation}
    T_i = f_{cluster}(X)
\end{equation}

Once the $T$ is extracted, the $s_{i,j}$ between the classes can be computed by taking the cosine distance (as per equation 3) between the feature centroids of each class. The reason for using cosine distance is that it achieves the highest mMAP score in CBIR systems when used with the feature vector of the average pooling layer. Finally, the matrix is generated by computing the pairwise distances of all classes with all other classes in the dataset. 

\begin{equation}
\operatorname{s_{i,j}}=\frac{T_i \cdot T_j}{\|T_i\| \times\|T_j\|}=\frac{\sum_{i=1}^{k} T_{i,k} \times T_{j,k}}{\sqrt{\sum_{k=1}^{n} T_{i,k}^{2}} \times \sqrt{\sum_{k=1}^{n} T_{i,k}^{2}}}
\end{equation}

\begin{figure}[!t]
\centering
\includegraphics[width = 0.9\columnwidth,scale=1.5]{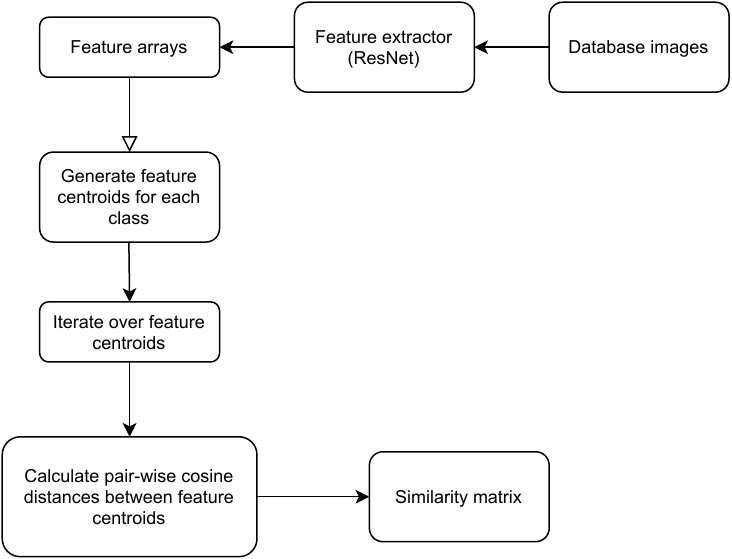}
\caption{Similarity matrix generation}
\label{flowchart}
\end{figure}

The synopsis of the entire methodology can be viewed in Figure \ref{flowchart}.

\subsection{Variety contribution ratio}\label{variety}

The variety contribution ratio ($v_r$) is the second value of the two-part DSI that can help in understanding the data, for a single class, in one go. It considers the overall spread of data compared to the number of samples ($n$) in the class cluster ($c$). We all know that one of the factors overfitting in an image-data algorithm is the variety of the dataset. Basically, the amount of data doesn’t always guarantee a better model, especially if the data samples are redundant. To check whether the data samples are "actually" contributing to the knowledge of the model or not, I define the variety contribution ratio. The contribution of the data samples is checked by finding the nearest feature $X_i$ to each sample and running a distance-based clustering algorithm like DBSCAN ($f_{db}$). The same features extracted earlier $X$ are passed to $f_{db}$. The distance can be set to cosine or any other metric but it has been noticed that the contribution ratio works well if the same distance metric is used for both similarity calculation and the variety contribution ratio, as both of them depend on which layer has been used to extract the feature vectors. 

\begin{equation}
    v = \rotatecharone{E}!(f_{db}(X))     
\end{equation}

The variety is calculated by the number of unique clusters after passing $X$ to $f_{db}$ with $min{\_}samples$ set to 1. If there is no other feature near(in the similarity distance threshold $z$) then the cluster will remain with a single sample or else it will contain more samples. $z$ can be set according to $S$ or a predefined context but a general value that is seen in most cases sets to about 0.05 (while using cosine distance). 

\begin{equation}
    v_r^{(i)} = \frac{v^{(i)}}{n} = \frac{\rotatecharone{E}!(f_{db}(X_i))}{n}
\end{equation}

If $v_r$ is 1, that means all the samples are actually contributing to the knowledge of the model, rather than contributing to overfitting or biasing. If $v_r$ is lower, like 0.7, then that means only 70 percent of the data is contributing and the other is not, hence can be comfortably removed to avoid unwanted bias due to data sample redundancy.

\begin{figure}[!t]
\centering
\includegraphics[width = 0.9\columnwidth,scale=1.5]{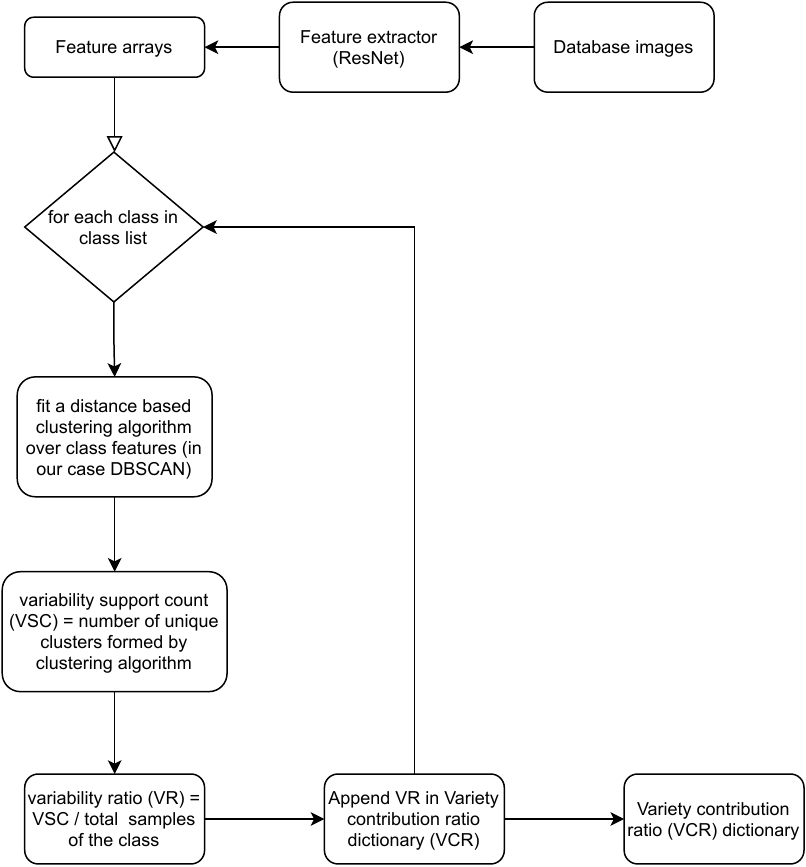}
\caption{Variety contribution ratio calculation}
\label{flowchart2}
\end{figure}

The synopsis of the entire methodology can be viewed in Figure \ref{flowchart2}.

\subsection{Model architectures}

I have used 4 traditional vision model architectures to train upon all the datasets. The architectures are: VGG-19, ResNet50, AlexNet and ResNeXt50. All the models are pre-trained on ImageNet and use the standard configuration with Cross entropy loss and step learning rate scheduler.

\subsection{Training environment }
All the models were trained on Intel Xeon E5-1620 v3 (@ 3.50GHz × 8 CPU)  with Nvidia GeForce RTX2080Ti GPU and 16GB of RAM. I used the standard PyTorch library to code and train the model architectures. All models were trained with CUDA enabled. \footnote{All code can be accessed at-https://github.com/Dishant-P/Dataset-structural-index-Official-research-module.git}

\begin{algorithm}
\textbf{INPUT:} $f{\_}list (n_{c} \times n_{i} \times 2048)$: List of feature arrays of every class image in the database\\
$class{\_}list$: List of all the classes in the dataset\\
\textbf{OUTPUT:} $sim{\_}mat (n_{c}\times n_{c})$: Similarity matrix\\
$n_{i}$: Number of images in that particular class\\ 
$n_{c}:$ Number of classes

\begin{algorithmic}
\STATE \textbf{Function} Similarity{\_}matrix{\_}generator($f{\_}list$, $class{\_}list$):
\STATE \hspace{0.3175cm} $i \leftarrow 0$
\STATE \hspace{0.3175cm}\textbf{for} length of $f{\_}list$ do

\STATE \hspace{0.635cm}$class{\_}features \leftarrow features[class{\_}list[i]]$ 
\STATE \hspace{0.635cm}$model \leftarrow KMeans(n{\_}clusters = 1)$
\STATE \hspace{0.635cm}$model.fit(class{\_}features)$
\STATE \hspace{0.635cm}$feature{\_}centroids[class{\_}list[i]] \leftarrow model.cluster{\_}centers{\_}$
\STATE \hspace{0.635cm}$i \leftarrow i + 1$

\STATE \hspace{0.5cm}
\STATE \hspace{0.3175cm} $i \leftarrow 0$
\STATE \hspace{0.3175cm}\textbf{for} length of $feature{\_}centroids$ do
\STATE \hspace{0.635cm} $j \leftarrow 0$
\STATE \hspace{0.635cm}\textbf{for} length of $feature{\_}centroids$ do
\STATE \hspace{0.9525cm}$simmat[class{\_}list[i]] \leftarrow cosine(feature{\_}centroids[i],$
\STATE \hspace{5.27cm}$feature{\_}centroids[j])$
\STATE \hspace{0.9525cm}$j \leftarrow j + 1$
\STATE \hspace{0.635cm}$i \leftarrow i + 1$
\STATE \hspace{0.3175cm}\textbf{return} $simmat$

\end{algorithmic}
\caption{Algorithm to generate Similarity Matrix}
\label{algo:algo1p1}
\end{algorithm}


\begin{algorithm}
\textbf{INPUT:} $f{\_}list (n_{c} \times n_{i} \times 2048)$: List of feature arrays of every class image in the database\\
$class{\_}list$: List of all the classes in the dataset\\
\textbf{OUTPUT:} $variety{\_}contribution{\_}ratio (n_{c}\times 1)$: Variety contribution ratio dictionary\\
$n_{i}$: Number of images in that particular class\\
$n_{c}:$ Number of classes

\begin{algorithmic}
\STATE \textbf{Function} calculate{\_}variability{\_}ratio($f{\_}list, class{\_}list$):
\STATE \hspace{0.3175cm}$variety{\_}contribution{\_}ratio \leftarrow \{ \}$
\STATE \hspace{0.3175cm}$i \leftarrow 0$
\STATE \hspace{0.3175cm}\textbf{for} length of $class{\_}list$ do: 
\STATE \hspace{0.635cm}$class{\_}features \leftarrow features[class{\_}list]$
\STATE \hspace{0.635cm}$X \leftarrow []$
\STATE \hspace{0.635cm}\textbf{for} $feature$ in $class{\-}features$:
\STATE \hspace{0.9525cm} $X.append(list(feature))$
\STATE \hspace{0.635cm}$model \leftarrow DBSCAN(eps=0.05, min{\_}samples=1,$
\STATE \hspace{5.27cm}$metric=`cosine`).fit(X)$
\STATE \hspace{0.635cm}$variability{\_}supporter{\_}count \leftarrow $ length of $set(model.labels{\_}) $
\STATE \hspace{0.635cm}$variability{\_}ratio \leftarrow variability{\_}supporter{\_}count / $ length of $X$
\STATE \hspace{0.635cm}$variety{\_}contribution{\_}ratio[class{\_}list[i]] \leftarrow variability{\_}ratio$
\STATE \hspace{0.635cm} $i \leftarrow i + 1$
\STATE \hspace{0.3175cm}\textbf{return} $variety{\_}contribution{\_}ratio$
\end{algorithmic}
\caption{Variety contribution ratio calculation}
\label{algo:algo2}
\end{algorithm}

\section{Experiments and Discussions}

\subsection{Variety contribution ratio usages}

\begin{enumerate}

    \item \textbf{Extract the maximum out of the current dataset and optimize further.} 
    
    The variety contribution ratio helps in reducing the class sample count without disturbing the validation accuracy of the class or the overall model. It means, with this, we can actually make an optimization over the dataset when we need to gain more data or are satisfied with the results of the data. It decreases the computational requirement for people with scarce resources. It also helps in improving the timings for the existing runs and reduces the time when retraining or adding additional visual capabilities. \\
    
    \begin{figure}[!t]
    \centering
    \includegraphics[width = 0.9\columnwidth,scale=1.0]{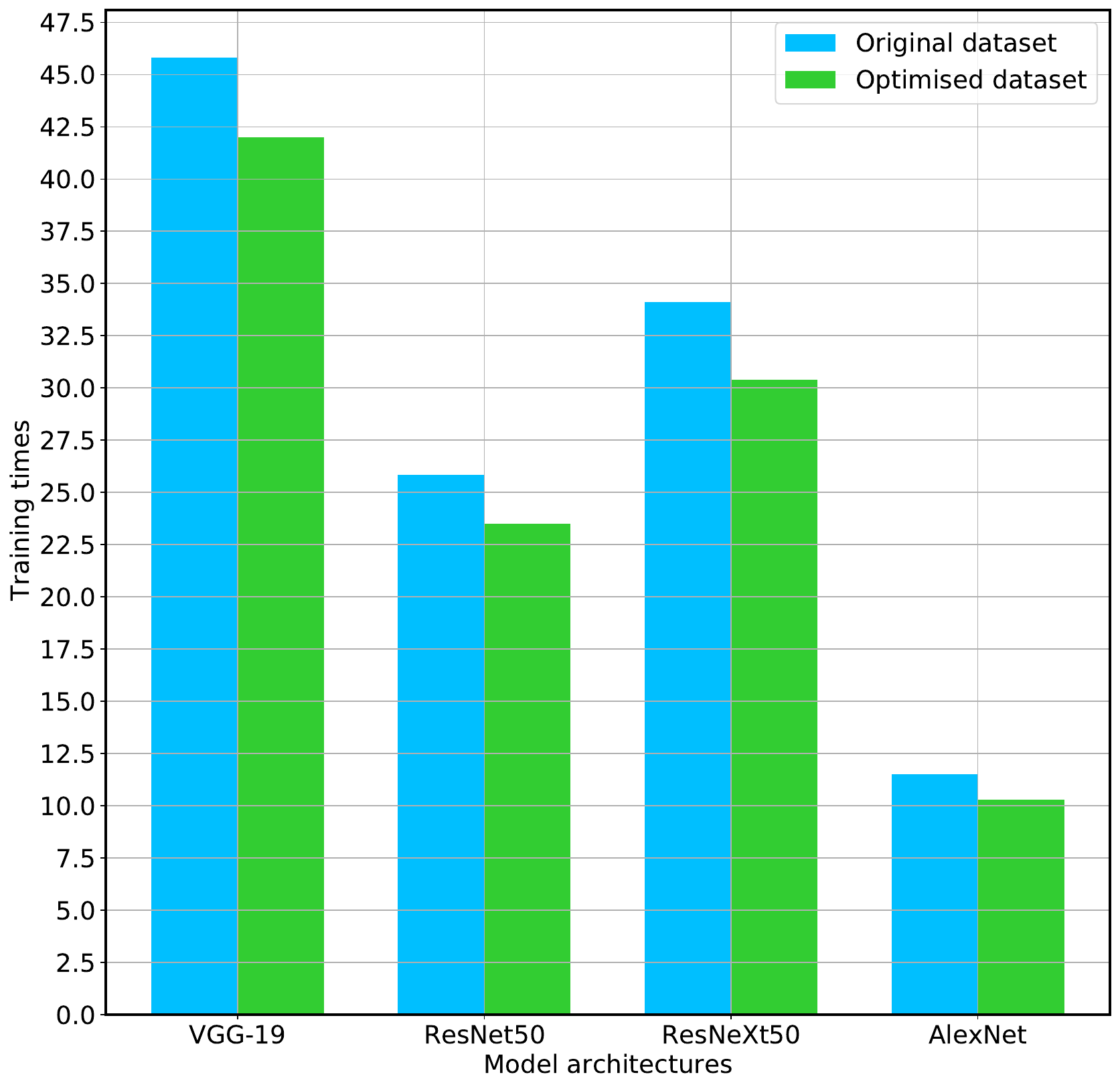}
    \caption{Comparing training times of each model architecture on original and optimised dataset.}
    \label{figure6}
    \end{figure}
    
    The variety contribution ratio can help in optimizing the existing dataset and reducing the number of resources required to train with the same knowledge, in turn, making room for more data(varied data) to be appended in the main dataset. Here I train three different model architectures namely, VGG-19, ResNet50, and ResNeXt50 on two versions of the Stanford Dogs dataset and Oxford flowers dataset, original and optimized dataset. The optimized dataset has not been changed in any manner only the number of samples in each class has decreased a little. I set the distance threshold for calculating the variety contribution ratio and remove the samples which are deemed redundant according to the Variety contribution ratio. The results of training on the full dataset and the optimized dataset can be viewed in Figure \ref{figure3} and \ref{figure7}.\\
    
    \begin{figure}[!t]
    \centering
    \includegraphics[width = 0.9\columnwidth,scale=1.0]{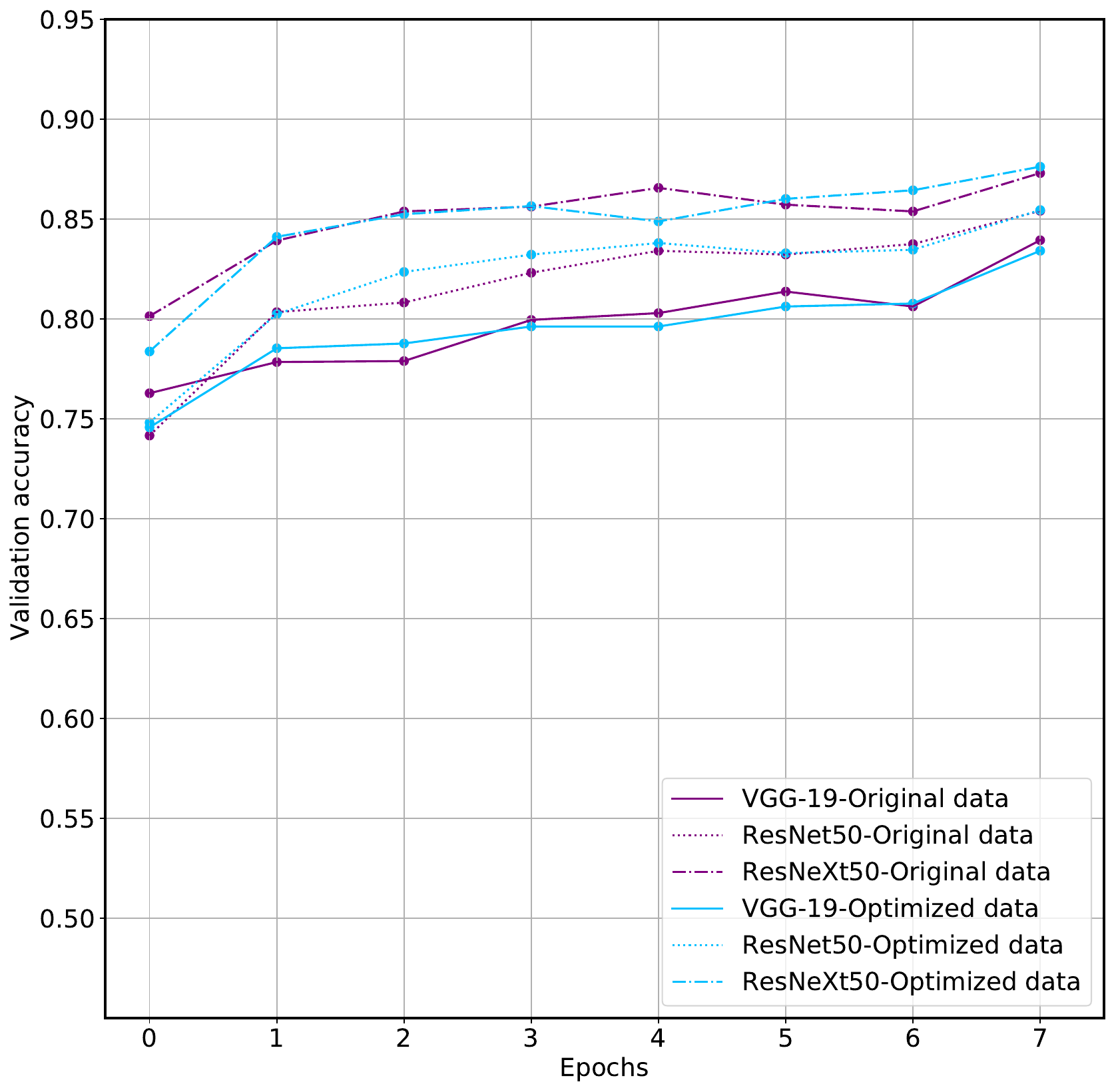}
    \caption{Validation accuracy on each epoch for Model architectures trained on original and optimized Stanford dogs dataset.}
    \label{figure3}
    \end{figure}
    
    \begin{figure}[!t]
    \centering
    \includegraphics[width = 0.9\columnwidth,scale=1.0]{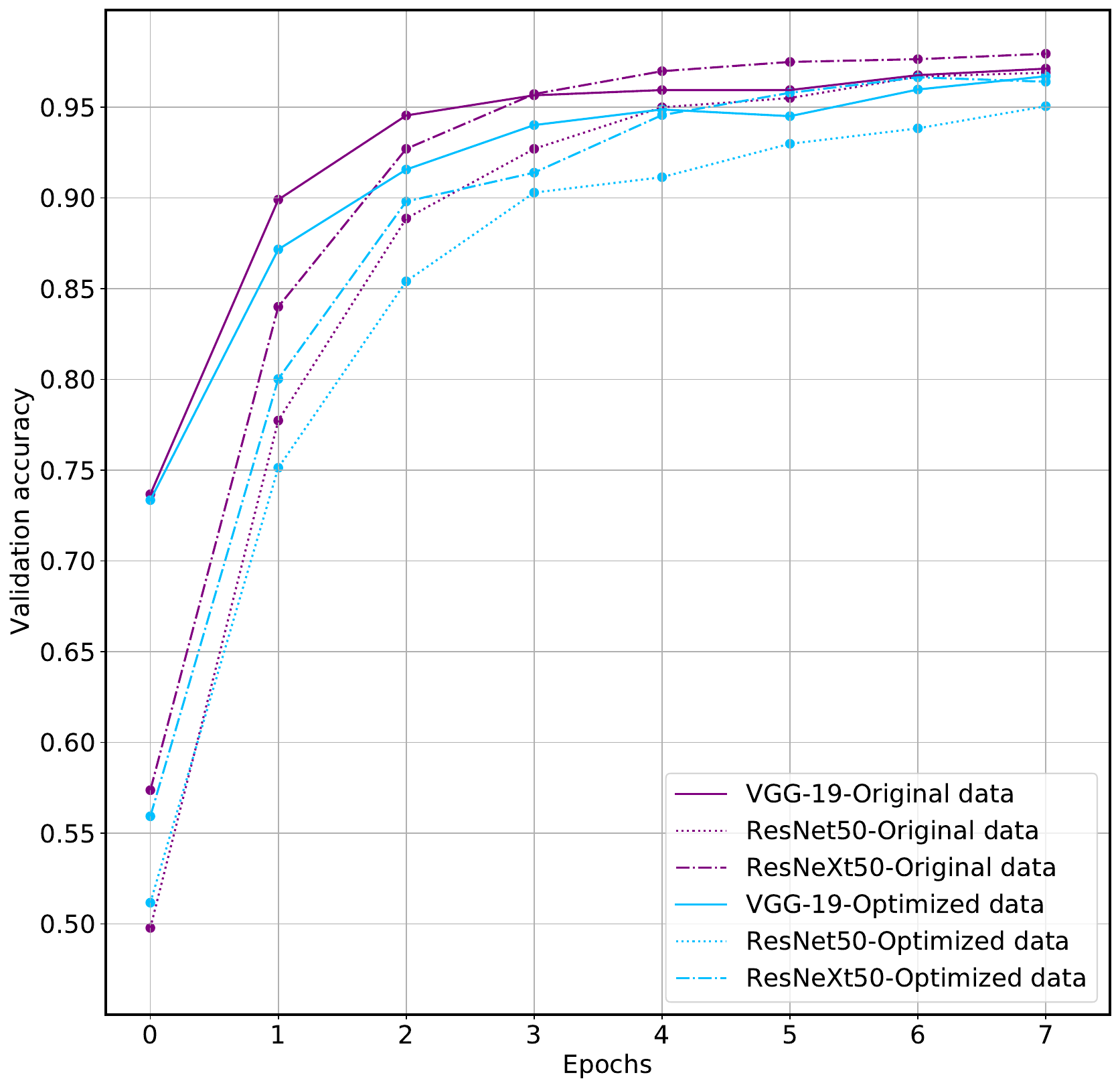}
    \caption{Validation accuracy on each epoch for Model architectures trained on original and optimized Oxford flowers dataset.}
    \label{figure7}
    \end{figure}
    
    In Figure \ref{figure6}, we can see that the models trained on the optimized dataset train 10{\%} faster than the ones trained on the original dataset. It shows how the variety contribution ratio can help save time in retraining due to reduced dataset size. Additionally, the model won't need the same amount of batches to complete an epoch as with the original dataset.
    
    In Figure \ref{figure3} and \ref{figure7}, I show the validation accuracies per epoch of all three model architectures trained on two versions of the same dataset. As can be seen, the validation accuracy of all three model architectures is nearly the same, with just a 0.5 percent deviation in the final accuracy. Not just that, but the entire training procedure looks nearly the same. On each epoch, the validation accuracy, as well as the training, can be seen going at the same pace in both the original and optimized version of the dataset. (Refer \ref{appendix:a} and \ref{appendix:b} for all class variety contribution ratio values.)

    \item \textbf{To figure out where the additional data is required.} 
    
    With the variety contribution ratio calculated on each class, it is much easier to figure out exactly which class needs further work in terms of getting more data or cleaning. It can help in justifying the class losses and corner one class at a time, and optimize its data. For example, if in a dataset of 5 classes, the class loss for one of the classes is too high, we now check not only the number of samples but also the variety contribution ratio and see if there is enough data according to the data samples used for training as well as testing. If not, more data could be collected to have more varied data samples fit by maintaining the variety contribution ratio after additional sample collection and reduce the overall class loss. With this, the collection of data could be more intuitive and more justifiable. 

\end{enumerate}

\subsection{Similarity matrix usages}

\begin{enumerate}
    \item \textbf{Using human-induced conditions to better highly similar classes classification.} 
    
    Once we can recognize the classes that the “machine” thinks are too similar, we can induce our human efforts to better the recognition by giving some of the obvious conditions to it. In Figure \ref{figure4}, there are two different species of flowers taken from the Oxford flowers dataset. One of them is a daffodil, and the second is a wildflower. The similarity or the cosine distance between them is 0.03 (which is highly similar). But if we look at them, they have a distinct characteristic (the overall color) that can be appended to the model’s decision which can make the classification between those two classes flawless. 
    
    \begin{figure}[!t]
    \centering
    \includegraphics[width = 0.9\columnwidth,scale=1.0]{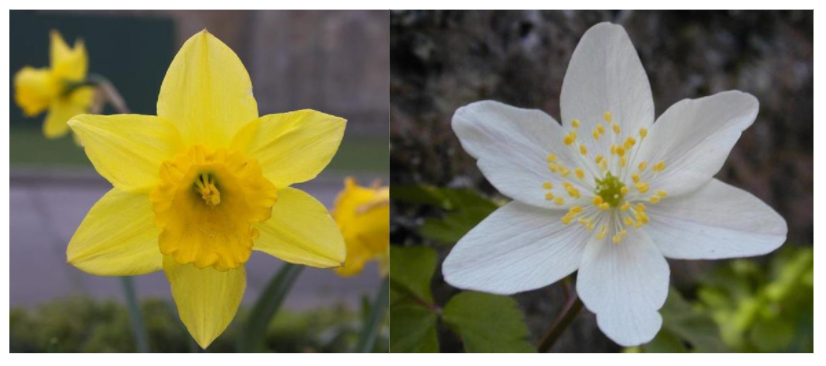}
    \caption{Left: Daffodil and Right: Wildflower}
    \label{figure4}
    \end{figure}

    \item \textbf{Understanding the balance between variety-contribution ratio and similarity matrix.}
    
    In any given dataset, there could be classes that are highly similar as well as highly distinct. The threshold for the variety contribution ratio can be adjusted according to the similarity of the class with other classes in the dataset. For example, if the class of reference is highly similar to other class(es) in the dataset, it is important to increase the variety contribution ratio for the class of reference and one or more classes that share high similarity with that class. As observed in the four-class dataset’s similarity matrix (Table \ref{table1}), all the classes are highly similar to each other, which can cause a lot of confusion for the model in terms of learning a proper classification between them. Hence, it is not advisable to compress the dataset too much with a very high similarity threshold in the variety contribution ratio calculation. Because with the similarity being this close, we would need samples very particular to that particular class and not discard them off as a redundant sample.

    \begin{table}[ht]
    \caption{Similarity matrix for four-class custom dataset.}
    \begin{center}
    \begin{tabularx}{\textwidth}{X X X X X}
    \hline
    \textbf{Class} & \textbf{Bus} & \textbf{Car} & \textbf{Truck} & \textbf{Van} \\ \hline
    Bus & 0 & 0.0977 & 0.0314 & 0.0468 \\
    Car & 0.977 & 0 & 0.0685 & 0.0378 \\
    Truck & 0.0314 & 0.0685 & 0 & 0.0292 \\
    Van & 0.0468 & 0.0378 & 0.0292 & 0 \\ \hline
    \end{tabularx}
    \label{table1}
    \end{center}
    \end{table}

    Alternatively, if the class is too distinct from other classes in the dataset, the variety contribution ratio need not be that high. Or the overall need for samples itself could be reduced. For example, in a dataset consisting of 5 classes, namely, ball, racket, watch, fish, and tiger, there is no need to indulge in collecting too many data samples or increasing the variety-contribution ratio. The work could easily be completed by less amount of images and normal variety contribution ratio as well. 
    
    \item \textbf{Making a better guess as to which model may better suit the dataset.} 
    
    The similarity matrix gives a crucial insight into the overall dataset, in terms of the complexity of the task at hand. If the task is classification or detection, the similarity matrix gives a hint at how difficult that classification could be for a particular model architecture. So, if the overall similarity is too high, shallow models may not work that well with a lower number of computations. Alternatively, if the similarity is too low, realistically speaking, simpler models could also work very well with a significantly lower number of computations. 
    
    For example, in a binary classification, if the similarity between the two classes is too low, there is no need to use neural networks to achieve that task. A simple SVM trained on the features could also do the task easily. Both would need to use CNNs for feature extractions, but SVM could be fitted way faster than a full CNN model. In figure \ref{figure5}, I show two classes from the Oxford flowers dataset, which have high dissimilarity of 0.188. One of the classes is a Pink Primrose, and the other is a Globe Thistle. By training a binary SVM classifier on the features extracted from the earlier method, I achieve 1.0 accuracy on the test set. 
    
    \begin{figure}[!t]
    \centering
    \includegraphics[width = 0.9\columnwidth,scale=1.0]{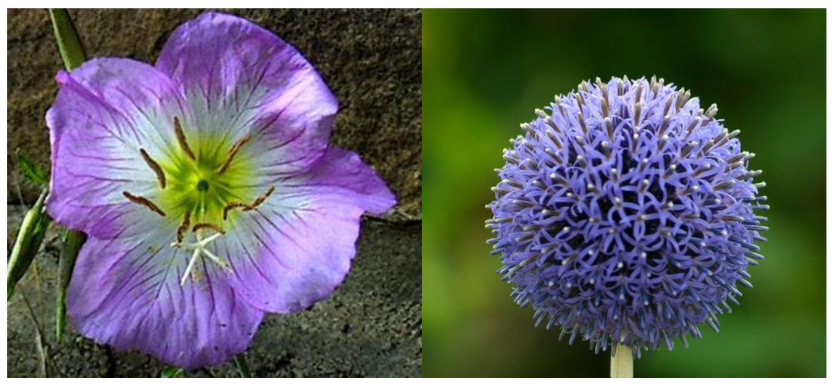}
    \caption{Left: Pink Primrose and Right: Globe Thistle}
    \label{figure5}
    \end{figure}

\end{enumerate}

\section{Future works and conclusion}
As with any new idea, there are a lot of improvements that are possible with DSI’s current methodology and architectures. There are three scenarios in which the work is still going on: First, the variety-contribution ratio’s threshold. Although the threshold could be set by experimenting, there still is the possibility to find out a proper value or mechanism to find the ideal threshold of VC ratio. Second, the way the distances are defined and the possibility of sub-vector comparisons. Right now, the cosine distance used to define the similarity value between two vectors averages the distances of each of the 2048 dimensions of the feature vector. But there could be a mechanism that effectively evaluates the dimensions on which the feature vector is highly similar and the part where it is distinct. This information could also be leveraged to improve the architecture and further understand the machine’s perspective. The last one is the ability to share the entire knowledge in one single value or equation to find a proper balance between the VC ratio and similarity values for each class. This can ease out the model selection techniques as well as dataset usability ranking creations, in turn making it easier for researchers and developers to select models and datasets for a particular task at hand.

I show how the dataset can be effectively evaluated using two of the measurements namely, variety contribution ratio and similarity matrix. The evaluation is shown individually, one for each class. I discussed some of the fundamental scenarios and applications of DSI. And the DSI need not be limited to just visual data. The same work that has been done on visual datasets could be performed on other types of data as well. I invite researchers from various fields to implement their architectures of the DSI on other data types. After a time there is going to be a data crisis and we need to make sure that we extract as much information from as little data as possible. Variety contribution factor is one such road to walk down.

\section{Funding Source Declaration}
This research did not receive any specific grant from funding agencies in the public, commercial, or not-for-profit sectors.

\bibliographystyle{elsarticle-num}
\bibliography{elsarticle-template}

\appendix \pagebreak
\section{Variety contribution ratio - Stanford Dog Breeds}
\label{appendix:a}

\begin{longtable}{llr}
\toprule
{} &                           Breed &       VCR \\
\midrule
1   &                       Chihuahua &  0.993421 \\
2   &                Japanese\_spaniel &  0.848649 \\
3   &                     Maltese\_dog &  0.976190 \\
4   &                        Pekinese &  0.979866 \\
5   &                             Tzu &  0.981308 \\
6   &                Blenheim\_spaniel &  0.989362 \\
7   &                        papillon &  0.923469 \\
8   &                     toy\_terrier &  0.994186 \\
9   &             Rhodesian\_ridgeback &  0.994186 \\
10  &                    Afghan\_hound &  0.958159 \\
11  &                          basset &  0.994286 \\
12  &                          beagle &  1.000000 \\
13  &                      bloodhound &  0.978610 \\
14  &                        bluetick &  0.976608 \\
15  &                   tan\_coonhound &  0.918239 \\
16  &                    Walker\_hound &  0.980392 \\
17  &                English\_foxhound &  0.923567 \\
18  &                         redbone &  0.993243 \\
19  &                          borzoi &  0.986755 \\
20  &                 Irish\_wolfhound &  0.986239 \\
21  &               Italian\_greyhound &  0.989011 \\
22  &                         whippet &  1.000000 \\
23  &                    Ibizan\_hound &  0.962766 \\
24  &              Norwegian\_elkhound &  0.897959 \\
25  &                      otterhound &  0.920530 \\
26  &                          Saluki &  0.955000 \\
27  &              Scottish\_deerhound &  0.905172 \\
28  &                      Weimaraner &  0.981250 \\
29  &       Staffordshire\_bullterrier &  0.987097 \\
30  &  American\_Staffordshire\_terrier &  0.993902 \\
31  &              Bedlington\_terrier &  0.846154 \\
32  &                  Border\_terrier &  0.994186 \\
33  &              Kerry\_blue\_terrier &  0.770950 \\
34  &                   Irish\_terrier &  0.982249 \\
35  &                 Norfolk\_terrier &  0.947674 \\
36  &                 Norwich\_terrier &  0.967568 \\
37  &               Yorkshire\_terrier &  1.000000 \\
38  &              haired\_fox\_terrier &  0.942675 \\
39  &                Lakeland\_terrier &  0.954315 \\
40  &                Sealyham\_terrier &  0.742574 \\
41  &                        Airedale &  0.995050 \\
42  &                           cairn &  1.000000 \\
43  &              Australian\_terrier &  0.974490 \\
44  &                  Dandie\_Dinmont &  0.977778 \\
45  &                     Boston\_bull &  0.989011 \\
46  &             miniature\_schnauzer &  0.993506 \\
47  &                 giant\_schnauzer &  0.974522 \\
48  &              standard\_schnauzer &  0.980645 \\
49  &                  Scotch\_terrier &  0.987342 \\
50  &                 Tibetan\_terrier &  0.941748 \\
51  &                   silky\_terrier &  0.972678 \\
52  &          coated\_wheaten\_terrier &  0.987179 \\
53  &     West\_Highland\_white\_terrier &  0.988166 \\
54  &                           Lhasa &  0.967742 \\
55  &                coated\_retriever &  0.980263 \\
56  &                coated\_retriever &  0.947020 \\
57  &                golden\_retriever &  1.000000 \\
58  &              Labrador\_retriever &  0.988304 \\
59  &        Chesapeake\_Bay\_retriever &  0.994012 \\
60  &                  haired\_pointer &  0.973684 \\
61  &                          vizsla &  0.993506 \\
62  &                  English\_setter &  0.987578 \\
63  &                    Irish\_setter &  0.993548 \\
64  &                   Gordon\_setter &  0.993464 \\
65  &                Brittany\_spaniel &  1.000000 \\
66  &                         clumber &  0.986667 \\
67  &                English\_springer &  0.943396 \\
68  &          Welsh\_springer\_spaniel &  0.980000 \\
69  &                  cocker\_spaniel &  0.993711 \\
70  &                  Sussex\_spaniel &  0.874172 \\
71  &             Irish\_water\_spaniel &  0.913333 \\
72  &                          kuvasz &  0.946667 \\
73  &                      schipperke &  0.961039 \\
74  &                     groenendael &  0.966667 \\
75  &                        malinois &  0.986667 \\
76  &                          briard &  0.940789 \\
77  &                          kelpie &  1.000000 \\
78  &                        komondor &  0.928571 \\
79  &            Old\_English\_sheepdog &  0.994083 \\
80  &               Shetland\_sheepdog &  0.993631 \\
81  &                          collie &  0.986928 \\
82  &                   Border\_collie &  0.993333 \\
83  &            Bouvier\_des\_Flandres &  0.980000 \\
84  &                      Rottweiler &  1.000000 \\
85  &                 German\_shepherd &  1.000000 \\
86  &                        Doberman &  1.000000 \\
87  &              miniature\_pinscher &  1.000000 \\
88  &      Greater\_Swiss\_Mountain\_dog &  0.928571 \\
89  &            Bernese\_mountain\_dog &  0.986239 \\
90  &                     Appenzeller &  0.980132 \\
91  &                     EntleBucher &  0.955446 \\
92  &                           boxer &  1.000000 \\
93  &                    bull\_mastiff &  0.980769 \\
94  &                 Tibetan\_mastiff &  0.960526 \\
95  &                  French\_bulldog &  1.000000 \\
96  &                      Great\_Dane &  0.987179 \\
97  &                   Saint\_Bernard &  0.970588 \\
98  &                      Eskimo\_dog &  0.993333 \\
99  &                        malamute &  1.000000 \\
100 &                  Siberian\_husky &  0.989583 \\
101 &                   affenpinscher &  0.953333 \\
102 &                         basenji &  0.985646 \\
103 &                             pug &  0.995000 \\
104 &                        Leonberg &  0.895238 \\
105 &                    Newfoundland &  0.994872 \\
106 &                  Great\_Pyrenees &  0.995305 \\
107 &                         Samoyed &  0.899083 \\
108 &                      Pomeranian &  0.986301 \\
109 &                            chow &  0.989796 \\
110 &                        keeshond &  0.911392 \\
111 &               Brabancon\_griffon &  0.960784 \\
112 &                        Pembroke &  0.988950 \\
113 &                        Cardigan &  0.993548 \\
114 &                      toy\_poodle &  1.000000 \\
115 &                miniature\_poodle &  0.993548 \\
116 &                 standard\_poodle &  0.993711 \\
117 &                Mexican\_hairless &  0.851613 \\
118 &                           dingo &  0.993590 \\
119 &                           dhole &  0.933333 \\
120 &             African\_hunting\_dog &  0.982249 \\
\bottomrule
\end{longtable}

\section{Variety contribution ratio - Oxford Flowers}
\label{appendix:b}
\begin{longtable}{llr}
\toprule
{} &             Flower species &       VCR \\
\midrule
1   &              Pink primrose &  1.000000 \\
2   &              Globe thistle &  0.921053 \\
3   &             Blanket flower &  0.885714 \\
4   &            Trumpet creeper &  0.938776 \\
5   &            Blackberry lily &  0.805556 \\
6   &                 Snapdragon &  0.852941 \\
7   &                Colt's foot &  0.753425 \\
8   &                King protea &  0.973684 \\
9   &              Spear thistle &  1.000000 \\
10  &                Yellow iris &  0.973684 \\
11  &               Globe-flower &  0.972222 \\
12  &          Purple coneflower &  0.850000 \\
13  &              Peruvian lily &  0.707692 \\
14  &             Balloon flower &  1.000000 \\
15  &  Hard-leaved pocket orchid &  0.877551 \\
16  &      Giant white arum lily &  1.000000 \\
17  &                  Fire lily &  0.941176 \\
18  &          Pincushion flower &  0.914894 \\
19  &                 Fritillary &  0.902778 \\
20  &                 Red ginger &  1.000000 \\
21  &             Grape hyacinth &  0.941176 \\
22  &                 Corn poppy &  0.969697 \\
23  &   Prince of wales feathers &  0.361111 \\
24  &           Stemless gentian &  1.000000 \\
25  &                  Artichoke &  0.951613 \\
26  &           Canterbury bells &  0.777778 \\
27  &              Sweet william &  0.590164 \\
28  &                  Carnation &  0.937500 \\
29  &               Garden phlox &  0.972222 \\
30  &           Love in the mist &  0.903226 \\
31  &              Mexican aster &  0.821429 \\
32  &           Alpine sea holly &  0.696970 \\
33  &       Ruby-lipped cattleya &  0.919355 \\
34  &                Cape flower &  0.391304 \\
35  &           Great masterwort &  0.840909 \\
36  &                 Siam tulip &  0.939394 \\
37  &                  Sweet pea &  0.909091 \\
38  &                Lenten rose &  0.944444 \\
39  &             Barbeton daisy &  0.505155 \\
40  &                   Daffodil &  0.979592 \\
41  &                 Sword lily &  0.960000 \\
42  &                 Poinsettia &  0.780822 \\
43  &           Bolero deep blue &  0.969697 \\
44  &                 Wallflower &  0.114650 \\
45  &                   Marigold &  0.344262 \\
46  &                  Buttercup &  0.736842 \\
47  &                Oxeye daisy &  0.578947 \\
48  &           English marigold &  0.740741 \\
49  &           Common dandelion &  0.643836 \\
50  &                    Petunia &  0.538835 \\
51  &                 Wild pansy &  0.522388 \\
52  &                    Primula &  0.800000 \\
53  &                  Sunflower &  0.914894 \\
54  &                Pelargonium &  0.375000 \\
55  &         Bishop of llandaff &  0.206522 \\
56  &                      Gaura &  0.700000 \\
57  &                   Geranium &  0.081395 \\
58  &              Orange dahlia &  0.214286 \\
59  &                 Tiger lily &  0.971429 \\
60  &         Pink-yellow dahlia &  0.035294 \\
61  &           Cautleya spicata &  0.305556 \\
62  &           Japanese anemone &  0.645833 \\
63  &           Black-eyed susan &  0.119048 \\
64  &                 Silverbush &  0.119048 \\
65  &          Californian poppy &  0.295455 \\
66  &               Osteospermum &  0.039216 \\
67  &              Spring crocus &  1.000000 \\
68  &               Bearded iris &  0.976744 \\
69  &                 Windflower &  0.717391 \\
70  &                Moon orchid &  0.939394 \\
71  &                 Tree poppy &  0.901961 \\
72  &                    Gazania &  0.859375 \\
73  &                     Azalea &  0.818182 \\
74  &                 Water lily &  0.721088 \\
75  &                       Rose &  0.830986 \\
76  &                Thorn apple &  0.863158 \\
77  &              Morning glory &  0.963855 \\
78  &             Passion flower &  0.795122 \\
79  &                      Lotus &  0.535714 \\
80  &                  Toad lily &  0.823529 \\
81  &           Bird of paradise &  0.971429 \\
82  &                  Anthurium &  0.987805 \\
83  &                 Frangipani &  0.600000 \\
84  &                   Clematis &  0.987805 \\
85  &                   Hibiscus &  0.836538 \\
86  &                  Columbine &  0.848485 \\
87  &                Desert-rose &  0.833333 \\
88  &                Tree mallow &  0.562500 \\
89  &                   Magnolia &  0.901961 \\
90  &                   Cyclamen &  0.931034 \\
91  &                 Watercress &  0.849673 \\
92  &                  Monkshood &  0.829268 \\
93  &                 Canna lily &  0.969697 \\
94  &                Hippeastrum &  0.728814 \\
95  &                   Bee balm &  0.698113 \\
96  &                  Ball moss &  0.970588 \\
97  &                   Foxglove &  0.946970 \\
98  &              Bougainvillea &  0.940594 \\
99  &                   Camellia &  0.930556 \\
100 &                     Mallow &  0.981481 \\
101 &            Mexican petunia &  0.926471 \\
102 &                   Bromelia &  0.880000 \\
\bottomrule
\end{longtable}

\end{document}